\documentclass[acmsmall]{acmart}
\usepackage[utf8]{inputenc}
\usepackage[T1]{fontenc}
\usepackage{amsmath}
\usepackage{dsfont}
\usepackage{soul,color}

\def\BibTeX{{\rm B\kern-.05em{\sc i\kern-.025em b}\kern-.08emT\kern-.1667em\lower.7ex\hbox{E}\kern-.125emX}}
    
\copyrightyear{2024}
\acmYear{2024}
\setcopyright{acmlicensed}
\acmConference[Woodstock '18]{Woodstock '18: ACM Symposium on Neural Gaze Detection}{June 03--05, 2018}{Woodstock, NY}
\acmBooktitle{Woodstock '18: ACM Symposium on Neural Gaze Detection, June 03--05, 2018, Woodstock, NY}

\begin{document}

%
\title{Gradient based Feature Attribution in Explainable AI: A Technical Review}

%

\author{Yongjie Wang}
\affiliation{\institution{Nanyang technological university}}
\email{yongjie002@e.ntu.edu.sg}

\author{Tong Zhang}
\affiliation{\institution{Nanyang technological university}}
\email{tong.zhang@ntu.edu.sg}

\author{Xu Guo}
\affiliation{\institution{Nanyang technological university}}
\email{xu.guo@ntu.edu.sg}

\author{Zhiqi Shen}
\affiliation{\institution{Nanyang technological university}}
\email{zqshen@ntu.edu.sg}

%
\renewcommand{\shortauthors}{Yongjie wang, et al.}

%
\begin{abstract}
The surge in black-box AI models has prompted the need to explain the internal mechanism and justify their reliability, especially in high-stakes applications, such as healthcare and autonomous driving. Due to the lack of a rigorous definition of explainable AI (XAI), a plethora of research related to explainability, interpretability, and transparency has been developed to explain and analyze the model from various perspectives. Consequently, with an exhaustive list of papers, it becomes challenging to have a comprehensive overview of XAI research from all aspects. Considering the popularity of neural networks in AI research, we narrow our focus to a specific area of XAI research: gradient based explanations, which can be directly adopted for neural network models. In this review, we systematically explore gradient based explanation methods to date and introduce a novel taxonomy to categorize them into four distinct classes. Then, we present the essence of technique details in chronological order and underscore the evolution of algorithms. Next, we introduce both human and quantitative evaluations to measure algorithm performance. More importantly, we demonstrate the general challenges in XAI and specific challenges in gradient based explanations. We hope that this survey can help researchers understand state-of-the-art progress and their corresponding disadvantages, which could spark their interest in addressing these issues in future work.
\end{abstract}

%
%

\ccsdesc[500]{Computing methodologies~Philosophical/theoretical foundations of artificial intelligence}
\ccsdesc[500]{Computing methodologies~Neural networks}
\ccsdesc[500]{Computing methodologies~Knowledge representation and reasoning}
\ccsdesc[500]{Theory of computation~Design and analysis of algorithms}

%
\keywords{Gradients, Feature attribution, Integrated Gradients, Explainable AI, Explanability}

\maketitle

\section{Introduction}

Nowadays, we are witnessing a remarkable surge of neural network models across various fields, e.g., computer version \cite{he2016deep,krizhevsky2012imagenet,liu2022swin}, natural language processing \cite{brown2020language,liu2019roberta, zhang-etal-2022-history}, robotics \cite{bousmalis2023robocat,lenz2015deep}, and healthcare \cite{jumper2021highly,senior2020improved}. Due to their opaque decision-making processes, AI models may exhibit biases toward ethnic minorities or make unexpected and potentially catastrophic errors. For example, ProPublica reported that the COMPAS justice system is biased towards African-American defendants by predicting a higher likelihood of their reoffending \cite{larson-2016}. \citeauthor{Ribeiro2016} \cite{Ribeiro2016} observed that the model discriminates between wolves and husky dogs with the existence of snow in the background. Therefore, there is an urgent need to elucidate the inner processes, understand the decision-making mechanisms, and enhance user trust of AI systems. 

\begin{figure}[tb]
    \centering
    \includegraphics[width=0.8\linewidth]{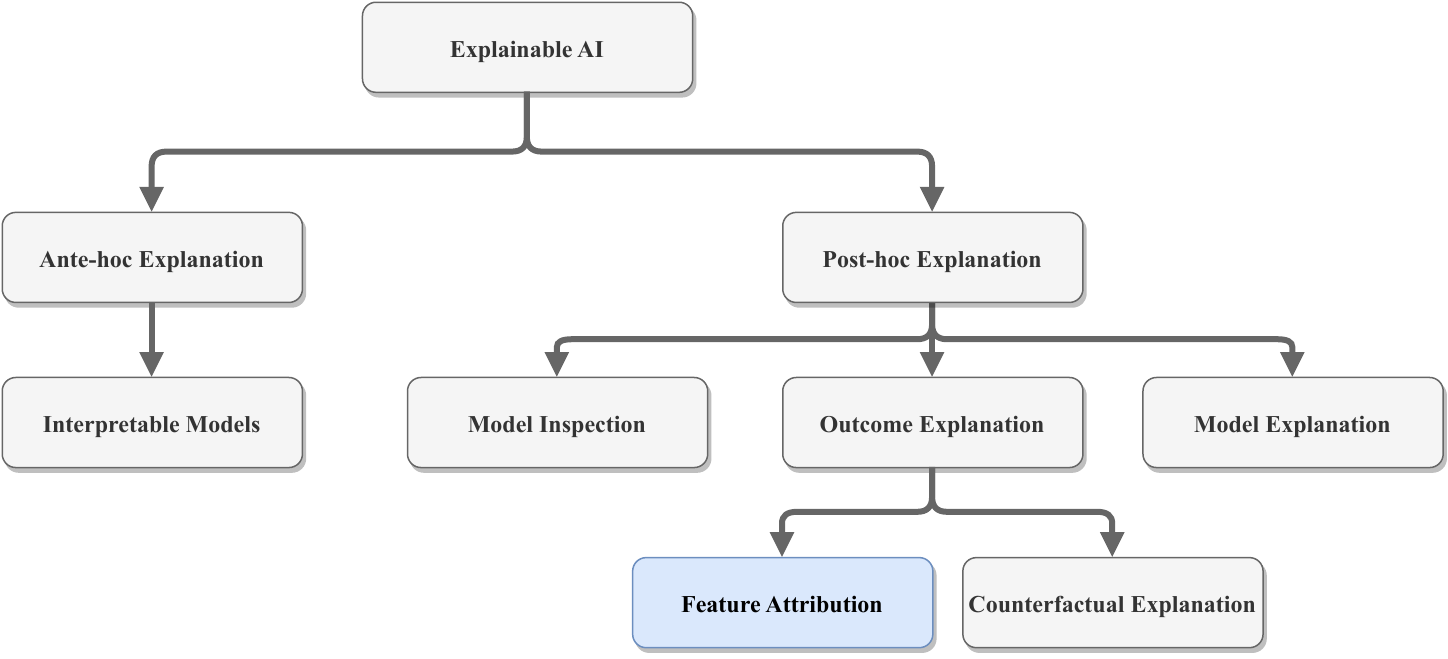}
    \caption{Taxonomy of Explainable AI according to \cite{guidotti2018survey}. In this research, we focus on gradient based explanations in feature attribution.}
    \label{fig:taxonomy}
\end{figure}

Explainable AI (XAI) refers to a series of techniques designed to reason and understand model behaviors, provide insights to rectify model errors/biases, and ultimately enable users to accept and trust in model's prediction. Following the taxonomy by \citeauthor{guidotti2018survey} \cite{guidotti2018survey}, shown as Figure \ref{fig:taxonomy}, XAI can be categorized into the following aspects: \textit{ante-hoc explanation} and \textit{post-hoc explanation}. \textit{Ante-hoc explanation} strives to develop transparent models that users can directly comprehend without the need for additional explanation tools, e.g., decision tree \cite{quinlan1986induction} and decision rule \cite{holte1993very}. \textit{Post-hoc explanation} aims to explain a trained black-box model by exploiting the relationship between input features and model predictions. \textit{Post-hoc explanations} can be further classified into \textit{model explanation} \cite{dhurandhar2018improving,lakkaraju2019faithful}, \textit{outcome explanation} \cite{Ribeiro2016,sundararajan2017axiomatic}, and \textit{model inspection} \cite{friedman2001greedy,goldstein2015peeking}. \textit{Model explanation} involves approximating the overall logic of a black-box model using an interpretable and transparent model at a global level. \textit{Outcome explanation} focuses on exploring the underlying reasons for a specific prediction at a local level. 
\textit{Model inspection} aims to offer visual and textual representations to facilitate understanding of the working mechanism of the model. 

Two approaches are commonly employed in outcome explanation: feature attribution (also referred to as feature importance approach) and counterfactual explanation. Feature attribution delves into directly identifying the importance of input features to the model's output while counterfactual explanation explores minimal and meaningful perturbations in the input space, to answer what changes in input values might affect the model's prediction. For a more in-depth exploration of the connections between both approaches, we refer readers to \citeauthor{kommiya2021towards}\cite{kommiya2021towards}. 

\subsection{Purpose of This Survey}

As there is no universal and rigorous definition of XAI, a plethora of research related to explainability, interpretability, transparency, and other related concepts fall within the XAI field. A search for the keywords ``explainable AI'' on Google Scholar yields more than $200,000$ results, presenting a formidable challenge to comprehensively elucidate all facets of XAI within a single publication. Although there have been many survey papers or book chapters \cite{adadi2018peeking, dovsilovic2018explainable, carvalho2019machine, tjoa2019survey, hohman2018visual, arrieta2020explainable, guidotti2018survey, gilpin2018explaining, dovsilovic2018explainable, samek2017explainable, liang2021explaining,molnar2020interpretable} on XAI, most of them give a brief description and demonstrate few early-stage works for a specific subfield of XAI such as gradient based feature attribution. The under-exploration of the specific subfield motivates us to comprehensively overview recent progress on gradient based explanations. Previous surveys aim to help practitioners quickly grasp various facets of XAI, whereas our survey paper delves into algorithmic details of gradient based explanation methods. By doing so, our objective is to assist researchers in applying the appropriate approach in more applications and fostering innovative breakthroughs in the narrow field. 

Based on different methodological approaches, \textit{feature attribution} contains the following branches of research: \textit{perturbation based methods} \cite{zeiler2014visualizing,fong2017interpretable,fong2019understanding}, \textit{surrogate based methods} \cite{Ribeiro2016,guidotti2018local}, \textit{decomposition based methods} \cite{bach2015pixel,binder2016layer,montavon2019layer,montavon2017explaining} and \textit{gradient-based methods} \cite{simonyan2013deep,sundararajan2017axiomatic,springenberg2014striving}. However, in this paper, our focus is on gradient-based methods, driven by the following considerations. 
\begin{itemize}
    \item \textbf{Instinct of gradients}. Gradients quantify how infinitesimal changes in input features impact the model predictions. Therefore, we can leverage gradients and their variants to effectively analyze the influence of feature modifications on the outcomes predicted by the model.
    
    \item \textbf{Seamless integration for neural networks}. Neural networks have gained great popularity and impressive performance across various domains. After model training, gradients can be readily obtained through a backward pass.  Therefore, gradient based explanations are straightforward to explain the neural networks without necessitating any changes to the models themselves. 
    \item \textbf{Satisfaction of axiomatic properties}. Due to the absence of ground truth, feature attribution methods may produce different explanations, leading to challenges in determining which one to trust.  Gradient based explanations are intentionally designed to satisfy certain axiomatic principles, such as sensitivity and completeness, ensuring to produce reasonable and desired explanations. 

\end{itemize}


\subsection{Our Contributions}

The contributions of our survey are summarized as follows:
\begin{itemize}
    \item We propose a novel taxonomy that systematically categorizes gradient based feature attribution into four groups. Subsequently, we introduce the gist of research motivation and technique detail for algorithms in each group.  
    
    
    \item We comprehensively overview a series of widely accepted evaluation metrics, including human evaluation and objective metrics, enabling a quantitative and qualitative comparison of the performance of various explanation methods. 
    
    \item We summarize both general research challenges in XAI and specific challenges unique to gradient based explanations, which could nourish and build a foundation for potential improvements in future work. 
\end{itemize}

\subsection{Research outline}
In the rest of this paper, Section \ref{sec:gradients} introduces philosophy and algorithm details of gradient based feature attribution; Section \ref{sec:evaluation} represents evaluation metrics of gradient based explanations, which is also applicable to other feature attribution methods; Section \ref{sec:challenges} describes the major concerns and challenges that may motivate further work in gradient based explanations; Section \ref{sec:conclusion} draws the conclusion. 
\section{Gradient based Feature Attribution}
\label{sec:gradients}

\begin{figure}[tbp]
    \centering
    \includegraphics[width=0.8\linewidth]{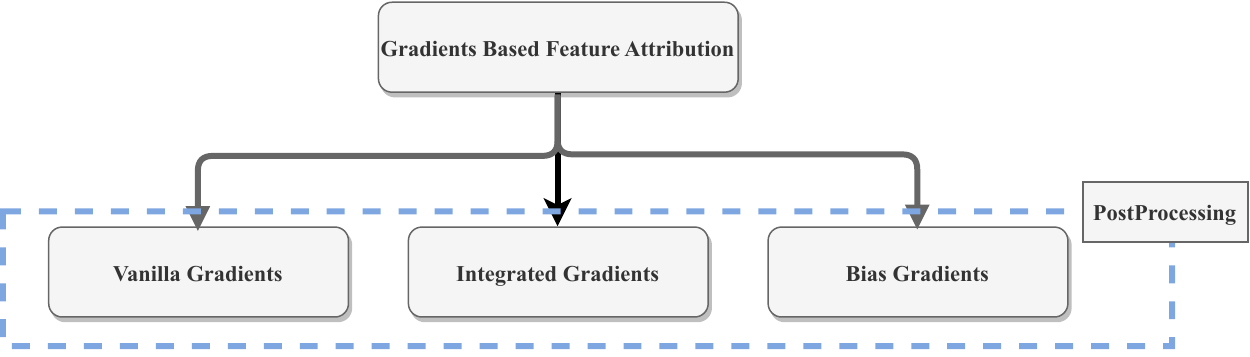}
    \caption{Taxonomy of gradient based feature attribution.}
    \label{fig:method}
\end{figure}

In this section, we classify gradients based feature attribution into four groups: \textit{vanilla gradients based explanation}, \textit{integrated gradients based explanation},\textit{ bias gradients based explanation}, and \textit{postprocessing for denoising}, as shown in Figure \ref{fig:method}. Vanilla gradients based explanation aims to explain model prediction with gradients (variants of gradients) through a backward pass. Integrated gradients based explanation refers to the methods that accumulate gradients along a path between a baseline point and the input. Bias gradients based explanation also considers the contribution of bias terms, and distributes the bias contribution to input features in conjunction with input gradients. At last, we discuss two post-processing methods that enhance the empirical quality of almost all gradients based explanations by adding noise to the input. Notations used in the following content are summarized in Table \ref{tab:notation}. 

\begin{table}[tbp]
\caption{Notation table.}
\label{tab:notation}
\begin{tabular}{|l|l|}
\hline
Notation & Meaning \\ \hline
$f(\cdot)$ & a trained neural network\\ \hline
$f_c(\cdot)$ & probability or logits of the class $c$\\ \hline
$\mathbf{x} \in \mathbb{R}^d$ & an input instance\\ \hline
$x_i$ & the $i$-th feature of $\mathbf{x}$ \\ \hline
$F(\cdot)$ & a predicted class, i.e., $F(\mathbf{x}) = \arg\max_{c} f_c(\mathbf{x})$\\ \hline
$\mathbf{a} \in \mathbb{R}^d$ & a feature importance vector of $\mathbf{x}$ \\ \hline
$a_i$ & the importance score of the $i$-th feature \\ \hline
$g(f, \cdot)$ or $g(\cdot)$ & a feature attribution method \\ \hline
$\hat{\mathbf{x}} \in \mathbb{R}^d$ & the baseline in integrated gradients based methods\\ \hline
$\gamma(\cdot)$ & the integral path in integrated gradients based methods\\ \hline
$W^l$ & weights of layer $l$ \\ \hline
$b^l$ & bias of layer $l$ \\ \hline
$z^l$ & intermediate feature of layer $l$ \\ \hline
$R^l$ & the relevance score of layer $l$ \\ \hline
$\phi(\cdot)$ & the activation function, e.g., ReLU  \\ \hline
\end{tabular}
\end{table}

\subsection{Preliminary}
For a dataset $\mathcal{D} = \{\mathbf{x}^{i}, \mathbf{y}^i\}, i \in \{1,2,\cdots,|\mathcal{D}|\}$, a neural network model is trained to minimize the empirical loss between $f(\mathbf{x}^i)$ and $\mathbf{y}^i$. Without loss of generality, we write a deep neural network model as,
\begin{align}
    f(\mathbf{x}) = W^m \phi (W^{m-1}\phi(...\phi(W^1 \mathbf{x} + b^1)...) + b^{m-1}) + b^m
    \label{eq:model}
\end{align}
where  $W^i$ and $b^i$  are trainable weights and bias term at layer $i$ and $\phi(\cdot)$ is a differential non-linear activation function. This formulation can generalize to most of the current feed-forward network models. For example, convolutional layers are essentially matrix multiplication after arranging the image into columns via the ``image to column'' operation; average pooling can be viewed as a linear transformation with the same weights.  

Due to the huge number of parameters in the model, which undermines the comprehension of the prediction logic, the objective of feature attribution methods is to discover the important features of $\mathbf{x} \in \mathbb{R}^d$ to the prediction, and their respective importance scores $a_i$. The objective is formally defined as,
\begin{definition}[Feature Attribution]
Given a black-box model $f: \mathcal{X} \subseteq \mathbb{R}^d \to \mathcal{Y}$ which makes a prediction $\mathbf{y} \in \mathcal{Y}$ for a given input $\mathbf{x} \in \mathcal{X}$, a feature attribution method $g: f \times \mathbb{R}^d \to \mathbb{R}^d$ calculates the importance scores $g(f, \mathbf{x})$ for a given instance $\mathbf{x}$, which measures the contribution/relevance of the instance $\mathbf{x}$'s feature values to the prediction $f(\mathbf{x})$. 
\end{definition}
\noindent For simplicity, we write $g(f, \mathbf{x})$ by $g(\mathbf{x})$ without loss of clearance. Given an input $\mathbf{x} = [x_1, x_2, ..., x_d]$, a feature importance method returns a feature importance vector $\mathbf{a} = [a_1, a_2, ..., a_d]$, where $\mathbf{a}$ help understand how the prediction $f(\mathbf{x})$ is made. In the domain of computer vision, the vector $\mathbf{a}$ can be visually represented through a heatmap, a.k.a. saliency map, effectively highlighting salient areas within the input image, which provides a clear and interpretable visualization for human evaluators. 


However, determining the optimal importance vector $\mathbf{a}$ is challenging and may even be impossible without imposing external constraints. On one hand, there are $d$ unknown variables and only one prediction, the underdetermined system have infinite number of solutions.  On the other hand, the definition of importance lacks rigorous mathematical support.  Generally, we define feature importance from a causal intuition, i.e., the removal of important features results in a sharp increase in the target prediction. Inevitably, the removal operation will bring unexpected noise and damage the prediction to some extent. It is challenging to disentangle the underlying reasons for the drop in prediction, whether stemming from the removal of important features or introduced noise. 

As calculating the importance vector is difficult, a commonly adopted approach \cite{sundararajan2017axiomatic,montavon2017explaining,shrikumar2017learning,binder2016layer} is to incorporate several prior axioms that a high-quality method should satisfy, which helps constrain the search for the importance vector. Need to note, these axioms, while necessary and desirable, may not be sufficient conditions to induce any meaningful and useful explanations if relying solely on them.  In the following, we introduce some widely accepted axioms.

\begin{itemize}
    \item \textit{Explainability}. It refers to the ability to provide insights into how a model arrives at a specific prediction or decision. Conventionally, we depict feature importance scores using a saliency map in computer vision and natural language processing, which highlights areas of interest for easy interpretation by humans. 
    
    \item \textit{Completeness} \cite{sundararajan2017axiomatic}, a.k.a., ``conservative'' in \cite{montavon2017explaining}, ``summation-to-delta'' in \cite{shrikumar2017learning}. The sum of attribution scores $\mathbf{a} \in \mathbb{R}^d$ should equal the prediction score $f(\mathbf{x})$ or the prediction difference between $f(\mathbf{x})$ and $f(\hat{\mathbf{x}})$, shown as follows,
    \begin{align}
    \label{eq:1}
        f(\mathbf{x}) = \sum_{i}^d a_i, \text{or\ },  f(\mathbf{x}) - f(\hat{\mathbf{x}}) = \sum_{i}^d a_i
    \end{align}
    
    \item \textit{Sensitivity} \cite{sundararajan2017axiomatic, shrikumar2017learning}. If modifying a feature results in a different prediction, that feature should have a non-zero attribution score. In addition, if a model is independent on a particular feature, the attribution score for that feature should be zero.
    
    \item \textit{Implementation invariance} \cite{sundararajan2017axiomatic}. Two functionally equivalent models(regardless of the internal implementation details) should have the same attribution results for the same instance. One can adopt the method outlined in \cite{kurkova1994functionally} and establish two functionally equivalent models to verify the satisfaction of this property.

    \item \textit{Linearity}. Suppose that a model $f$ can be linearly decomposed into two models $f_1$ and $f_2$, i.e.,  $f = w_1  f_1 + w_2  f_2$. It requires the attribution scores should preserve the same linearity, $g(f, \mathbf{x}) = w_1  g(f_1, \mathbf{x}) + w_2  g(f_1, \mathbf{x})$. 
    
    \item \textit{Symmetry preservation}. Two features are symmetric if swapping them does not change the prediction. If all inputs have identical values for symmetric features and baselines also have identical values for symmetric features, then an attribution method should assign the identical importance scores on symmetric features. 
    
\end{itemize}

\subsection{Vanilla Gradients based Explanation}

Gradients of the model prediction with respect to the input reflect how the prediction behaves in response to infinitesimal changes in inputs. As gradients serve as a local approximation of the feature coefficients, directly using gradients for explaining the model is a reasonable starting point. Here, we refer to these explanations simply derived from gradients as ``vanilla gradients based explanations.''

\textbf{Backpropagation}\cite{simonyan2013deep}. \citeauthor{simonyan2013deep} proposed an image-specific class saliency visualization. For a given image $\mathbf{x}_0$, the model returns a prediction score $f_c(\mathbf{x}_0)$ of a class $c$. \textit{Backpropagation} uses the first-order Taylor expansion to approximate the model $f$ in the neighbourhood of $\mathbf{x}_0$,
\begin{align}
    f_c(\mathbf{x}) = w^T \mathbf{x} + b
\end{align}
$w^T$ can be viewed as the local feature coefficient on the prediction $f_c(\mathbf{x})$, specifically, 
\begin{align}
    w = \frac{\partial f_c(\mathbf{x})}{\partial \mathbf{x}} \bigg| _{\mathbf{x}_0}
\end{align}
For a pixel $(i,j)$ with Red, Green, and Blue (R, G, B) channels, it takes the maximum of the absolute values for each pixel, i.e., $max_c|w_{(i,j,c)}|$. Similarly, \citeauthor{gevrey2003review} \cite{gevrey2003review} also propose to consider both the gradients and the square of gradients to explain the model. 

\textbf{Deconvolutional network (deconvnet)} \cite{zeiler2014visualizing}. Deconvnet starts with an activation at a specified layer and maps the activation back to the input space, to demonstrate what input pattern activates this neuron. During activation inversion, it replaces the max pooling layer in the forward pass with the unpooling operation by recording the locations of maxima, the convolution layer with a deconvolution layer by transposing the filters, and keeps the rectification units (ReLU) to merely allow the positive signal to pass.  To eliminate the interference of other neurons in that layer, it first sets all other activations to zero, then iteratively passes the modified feature map to the replaced deconvnet layers until it reaches the input space. A general framework that generalizes to backpropagation and deconvnet can be found in SaliNet \cite{mahendran2016salient}.

\textbf{Guided backpropagation} \cite{springenberg2014striving}. \citeauthor{springenberg2014striving} introduced a straightforward network architecture that substitutes the max pooling layer with a convolutional layer (with stride) without compromising accuracy. Considering ``deconvnet'' does not perform well without max-pooling layers, they proposed a novel visualization method, named \textit{guided backpropagation}. When propagating through the Rectified Linear Units (ReLUs), it zeros out the entries whose values are negative in both the forward pass and backward pass. 

\textbf{Rectified Gradient (RectGrad)} \cite{kim2019saliency}. \citeauthor{kim2019saliency} hypothesize that the noise may originate from irrelevant features have positive pre-activated values when they pass through the ReLUs . When we compute gradients with backpropagation, these unrelated features have non-zero gradients. Consequently, they propose a thresholding method that enables gradients to propagate through ReLUs only if their importance scores surpass a specified threshold. Specifically, 
\begin{align}
    R_i^{(l)} = (z_i^{(l)} * R_i^{(l+1)} > \tau) \cdot R_{i}^{(l+1)}
\end{align}
Here, $z_i^{(l)}$ represents the activated value after ReLU, and $\tau$ is the threshold set to the $q^{th}$ percentile of each entry importance at layer $l$. 

\begin{figure}
    \centering
    \includegraphics[width=\textwidth]{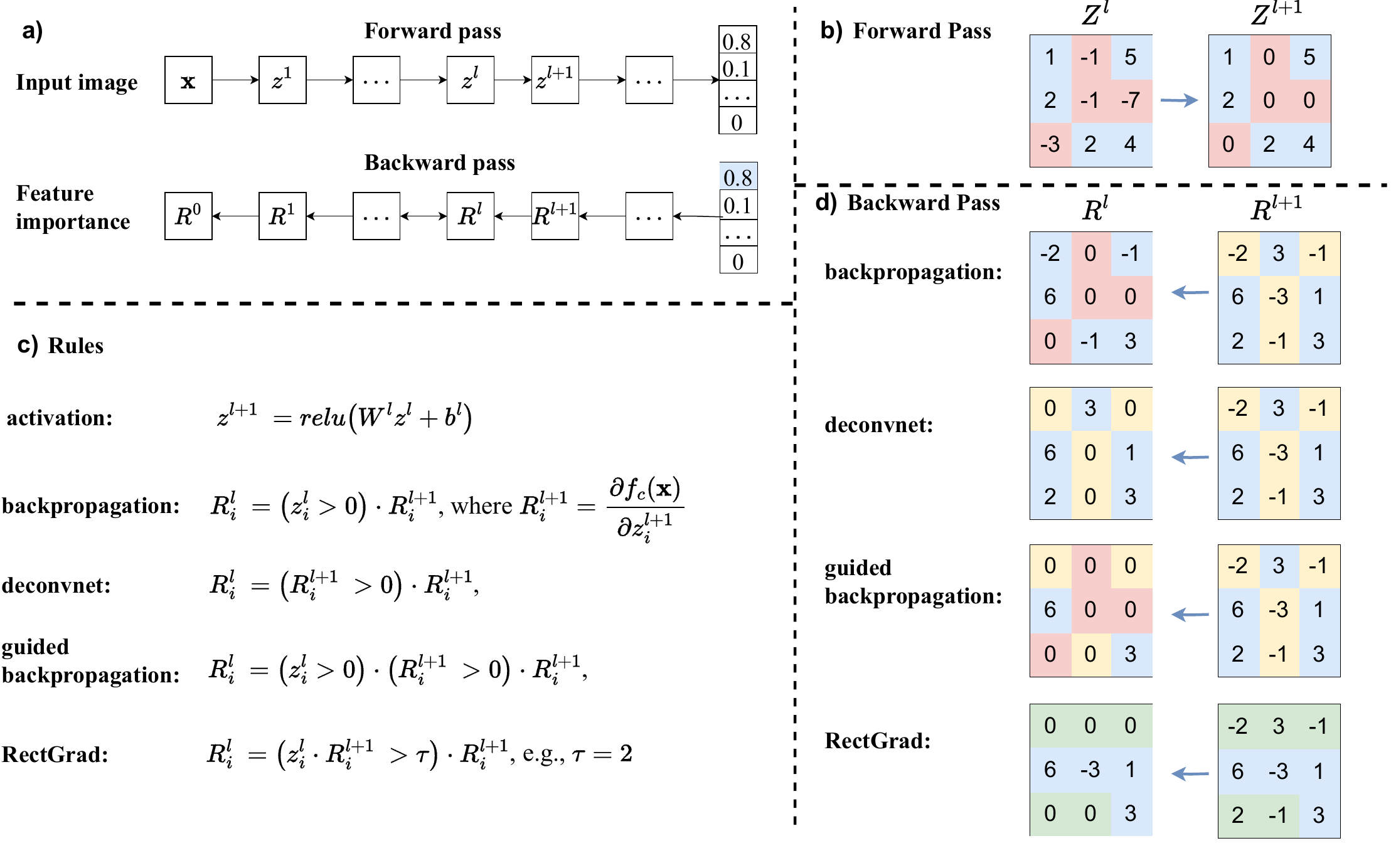}
    \caption{The differences among \textit{backpropagation}, \textit{deconvolutional network}, \textit{guided backpropagation}, and \textit{RectGrad} lie in the implementation of ReLU in the backward pass.}
    \label{fig:comparison}
    \vspace{-0.5cm}
\end{figure}

The comparison among the \textit{backpropagation}, \textit{deconvolutional network}, \textit{guided backpropagation} and \textit{RectGrad} is summarized in Figure \ref{fig:comparison}. The major difference lies in the ways to process backpropagation through the ReLU layer.  \textit{Backpropagation} uses the gradients w.r.t. the input instance, retaining the signal only when the inputs of ReLU are positive in the forward pass. \textit{Deconvnet} and \textit{Guided Backpropagation} employ a variant of gradients. \textit{Deconvnet} leverages ReLUs to filter out negative gradients during the backward pass and \textit{Guided Backpropagation} combines the strategies of \textit{Backpropagation} and \textit{Deconvnet} simultaneously. \textit{RectGrad} filters out certain entries whose values are below a threshold. A clear drawback of \textit{Deconvnet} and \textit{Guided Backpropagation} is their inability to identify the negative influence on the output. 

\textbf{Gradient $\times$ Input} \cite{ancona2017towards,shrikumar2017learning}. Layer-wise Relevance Propagation (LRP) \cite{bach2015pixel} computes the relevance score by redistributing the prediction score by some predefined rules layer by layer. 
However, researchers \cite{ancona2017towards,shrikumar2017learning} have observed that $\epsilon$-LRP is equivalent, within a scaling factor, to a feature-wise product between the input and its gradients, if numerical stability is not considered. 

\textbf{Gradient-weighted Class Activation Mapping (Grad-CAM)} \cite{selvaraju2017grad}. 
Prior studies \cite{oquab2015object, zhou2014object} have shown that CNNs can proficiently localize objects of interest solely under the supervision of image-level labels. Class Activation Mapping (CAM), as introduced by \cite{zhou2016learning}, is a technique to identify discriminative regions that the model use to make a prediction, through the linearly weighted summation of activation maps from the last convolution layer. It can be formulated under the following equation,
\begin{align}
    g^{CAM}(\mathbf{x}) = ReLU(\sum_{k=1}^K \omega_k A_k)
\end{align}
where $A_k$ is $k$-th channel of the activation map of last convolution layer, which can be obtained in a forward pass, and $\omega_k$ is the importance of each channel. Because CAM requires the explicit definition of a global pooling layer for computing $\omega_k$, Grad-CAM \cite{selvaraju2017grad} introduce a general method that uses the gradients w.r.t. the activation map $A_k$ to measure the channel importance, i.e.,
\begin{align}
    \omega_k =  GP(\frac{\partial f_c(\mathbf{x})}{\partial A_k})
\end{align}
Here, $GP(\cdot)$ is the global average pooling operator. Extending Grad-CAM, there are several methods proposed, e.g., Grad-CAM++ \cite{chattopadhay2018grad}, Smooth Grad-CAM++ \cite{omeiza2019smooth}. Since these methods directly utilize gradients and their variants to acquire channel importance of the last convolutional layer, we omit further description of them.

In addition to above methods, numerous research incorporates the vanilla gradients to domain-specific models or tasks \cite{Pope_2019_CVPR,li-etal-2016-visualizing}. As we mainly focus on algorithm details, the domain-specific applications are not in the scope of our discussion. Some explanation methods adopt vanilla gradients as sub-components within their algorithms (for example, Relative Sectional Propagation (RSP) \cite{nam2021interpreting} utilizes gradients w.r.t. the last convolution layer as the initialization of relevance propagation; SGLRP \cite{9022542} uses gradients of probability after softmax w.r.t. intermediate features to suppress relevance propagation of non-target classes). We also omit these methods because their primary contributions do not involve enhancing gradient based explanations. 

\subsection{Integrated Gradients based Explanation}

Gradients indicate the infinitesimal prediction changes resulted from infinitesimal feature changes. For an input $\mathbf{x}$ receiving the higher prediction, the tiny prediction change cannot reflect the reasons behind the prediction.  In addition, vanilla gradients  are susceptible to the saturation issue \cite{sturmfels2020visualizing,sundararajan2017axiomatic}, wherein the gradients of some features are close to $0$, even if the model heavily depends on those features. Therefore, researchers introduce the baseline (a.k.a., reference point) from the counterfactual philosophy, where the baseline represents the absence or neutral of the current prediction. Integrated gradients based explanations typically accumulate gradients along a specified path from a baseline to the input point.


\textbf{Integrated Gradients (IG)} \cite{sundararajan2017axiomatic}. \citeauthor{sundararajan2017axiomatic} firstly propose the integrated gradients that satisfies a number of desired properties, like sensitivity, completeness, linearity and implementation invariance. Mathematically, the gradient vectors of the model $f$ at all data points form the conservative vector field. According to the fundamental Gradient Theorem in line integrals \cite{lerma2021symmetry}, we have,
\begin{align}
    f(\gamma(1)) - f(\gamma(0)) &= \int_{\gamma} \bigtriangledown f(\mathbf{x})\cdot d\mathbf{x} \\
    &= \int_{\gamma} \sum_{i=1}^n \frac{\partial f(\mathbf{x})}{\partial x_i} d x_i \\
    &= \sum_{i=1}^n \int_{\gamma} \frac{\partial f(\mathbf{x})}{\partial x_i} d x_i
\end{align}
where $\gamma:[0, 1] \to R^d$ is the smooth path function, and $\bigtriangledown$ is the nabla operator, i.e., the gradients at a point $\mathbf{x}$. The authors define the IG score of the $i$-th feature of $\mathbf{x}$ by,
\begin{align}
    g^{IG}_i(\mathbf{x}) = \int_{\gamma} \frac{\partial f(\mathbf{x})}{\partial x_i} d x_i
    \label{eq:ig}
\end{align}
Obviously, the IG score of each feature $i$ heavily depends on the path function $\gamma$.  For simplicity, they choose the straight line from the baseline $\hat{\mathbf{x}}$ to input $\mathbf{x}$, where a point in the path can be written as, $\gamma(t) = \hat{\mathbf{x}} + t(\mathbf{x} - \hat{\mathbf{x}})$. Replacing $dx_i$ with $(x_i - \hat{x}_i)dt$, we can obtain,
\begin{align}
    g^{IG}_i(\mathbf{x}, \hat{\mathbf{x}}) = (x_i - \hat{x}_i) \int_{0}^1 \frac{\partial f(\gamma(t))}{\partial x_i} dt
    \approx (x_i - \hat{x}_i) \times \sum_{k=1}^m \frac{\partial f(\hat{\mathbf{x}} + \frac{k}{m}(\mathbf{x} - \hat{\mathbf{x}})) }{\partial x_i} \times \frac{1}{m}
    \label{eq:dig}
\end{align}
With this approach, the attribution score of each feature can be calculated by accumulating the gradients of samples along the path. If a straight line is adopted, IG is also symmetry-preserving. \citeauthor{hesse2021fast} \cite{hesse2021fast} demonstrated that for a nonnegatively homogeneous model, Integrated Gradients (IG) with a zero baseline is equivalent to Input$\times$Gradient \cite{shrikumar2017learning}. 

\textbf{Blur Integrated Gradients (BlurIG)} \cite{xu2020attribution}. BlurIG aims to explain the prediction by accumulating gradients from both the frequency and space domains. Specifically, the path of BlurIG is defined by successively blurring the input image with a Gaussian blur filter. Formally, let $\mathbf{x}(p, q)$ denotes a pixel at the location $(p, q)$, the discrete convolution of the input signal with the 2D Gaussian kernel with variance $\sigma$ can be written as,  
\begin{align}
    L(p,q,\sigma) = \sum_{m = -\infty}^{\infty} \sum_{n = -\infty}^{\infty} \frac{1}{\pi \sigma} e^{\frac{-(p^2+q^2)}{\sigma}} \mathbf{x}(p-m, q-n)
\end{align}
Then, the final attribution score at location $(p, q)$ of BlurIG is computed as, 
\begin{align}
    g^{BlurIG}(p,q) = \int_{\sigma = \infty}^0 \frac{\partial F(L(p,q,\sigma))}{\partial L(p,q,\sigma)} \frac{\partial L(p,q,\sigma)}{\partial \sigma} d\sigma 
    \label{eq:blurig}
\end{align}
In the limit as $\sigma$ approaches infinity, the maximally blurred image converges to the mean value across all locations. Setting $\sigma = 0$ corresponds to the input image itself. In this way, BlurIG avoids the explicit definition of baseline, which is difficult to choose in some applications. Furthermore, BlurIG has the potential to alleviate artifacts that may arise during interpolation when new features are introduced along the integral line. 

\textbf{Expected Gradients} \cite{sturmfels2020visualizing}.  The baseline in Integrated Gradients (IG) represents the absence of the target object, which can be challenging to explicitly define. By default, IG chooses the all-zero vector as the baseline. However, in cases where the body of the target object is black, IG fails to highlight the body area effectively. Furthermore, the value $0$ often has a unique meaning in tabular datasets and may not appropriately represent the concept of missingness. 

There are alternative choices for baselines in Integrated Gradients (IG), such as an image with the maximum distance from the current input, a Gaussian-blurred image, an image with random pixel values, and black \& white baseline \cite{kapishnikov2019xrai}. Nevertheless, each baseline choice has its advantages and disadvantages. A natural way is to average the attribution scores of multiple baselines that are sampled from a baseline distribution $D$. 
 \begin{align}
     g_i^{EG}(f, \mathbf{x}) = \int_{\hat{\mathbf{x}}} g_i^{IG}(\mathbf{x}, \hat{\mathbf{x}}) \times p_{D}(\hat{\mathbf{x}}) d \hat{\mathbf{x}}
 \end{align}
 here, $g^{IG}_i(\mathbf{x}, \hat{\mathbf{x}})$ denotes the attribution score of $i$-th feature from IG with a baseline $\hat{\mathbf{x}}$, $p_{D}(\cdot)$ is the density function of baseline distribution. 

The integral over the baseline distribution is approximated with Riemann integration. Specifically, expected gradients simply sum the gradients over $k$ samples from the $k$ straightline paths with the following formula,
\begin{align}
   g_i^{EG}(f, \mathbf{x}) = \frac{1}{k} \sum_{j=1}^k (x_i - \hat{x}^j_{i}) \times \frac{\partial f(\hat{\mathbf{x}}^j + \alpha_j(\mathbf{x}-\hat{\mathbf{x}}^j))}{\partial x_i}
\end{align}
where $\hat{\mathbf{x}}^j$ is the $j$-th baseline from baseline distribution $D$ and $\alpha_j\sim U(0,1)$ is sampled from the uniform distribution to determine the interpolated datapoint between $\mathbf{x}$ and $\hat{\mathbf{x}}$. 

\textbf{Split Integrated Gradients (Split IG)} \cite{miglani2020investigating}. In integrated gradients, the interpolated images along the path are determined by the scaling factor $\alpha$. A common phenomenon is that the model prediction sharply spikes with $\alpha$ and then plateaus as $\alpha$ scales up to $1$.  The range of $\alpha$ wherein the model prediction exhibits minimal variation is denoted as the saturated region. \citeauthor{miglani2020investigating} observed that the accumulated gradients within the saturation regions constitute a non-negligible contribution of the final importance scores of IG, even though these regions have little impact on improving the model prediction. As suggested by its name, \citeauthor{miglani2020investigating} split the integral of IG into two parts: the non-saturated region where the prediction increases substantially, and the saturation region, controlled by a hyperparameter $\varphi$,
\begin{align}
    \alpha^* = \text{inf}\{ \alpha \in [0, 1], \text{s.t.,} f(\hat{\mathbf{x}} + \alpha (\mathbf{x} - \hat{\mathbf{x}})) > f(\hat{\mathbf{x}}) + \varphi (f(\mathbf{x}) - f(\hat{\mathbf{x}})) \}
\end{align}
Based on their observation, Split IG merely integrates the gradient in the non-saturated region where $\alpha < \alpha^*$.

\textbf{Integrated Hessians} \cite{janizek2021explaining}. While much research has concentrated on elucidating the significance of individual features for the present prediction, the exploration of explaining feature interactions has received comparatively less attention. Integrated Hessians treat the integrated gradient method $g^{IG}(f, \mathbf{x})$ as a differential function, and explain the importance of feature $i$ in terms of the feature $j$ 
with second-order derivatives,
\begin{align}
    \Gamma_{i, j}(\mathbf{x}) = g_j^{IG}(g_i^{IG}(f, \mathbf{x}), \mathbf{x})
\end{align}
Here, we write the IG method by $g^{IG}(f, \mathbf{x})$, omitting the baseline notation to emphasize the differences in the differential function. For a differential model $f$, $\Gamma_{i, j}(\mathbf{x})$ has the following mathematical forms,
\begin{align}
    \Gamma_{i, j}(\mathbf{x}) &= (x_i - \hat{x}_i)(x_j - \hat{x}_j) \times \int_{\beta=0}^1 \int_{\alpha=0}^1 \alpha  \beta \frac{\partial^2 f(\hat{\mathbf{x}} + \alpha\beta(\mathbf{x}-\hat{\mathbf{x}}))}{\partial x_i \partial x_j} d\alpha d\beta, \quad\quad\quad\   i\neq j \\
    &= (x_i - \hat{x}_i)^2 \times \int_{\beta=0}^1 \int_{\alpha=0}^1 \alpha  \beta \frac{\partial^2 f(\hat{\mathbf{x}} + \alpha\beta(\mathbf{x}-\hat{\mathbf{x}}))}{\partial x_i \partial x_j} d\alpha d\beta  \qquad \qquad \qquad \quad i = j\\
    &+ (x_i - \hat{x}_i) \times \int_{\beta=0}^1 \int_{\alpha=0}^1 \frac{\partial f(\hat{\mathbf{x}} + \alpha\beta(\mathbf{x}-\hat{\mathbf{x}}))}{\partial x_i} d\alpha d\beta \nonumber
\end{align}
The Riemann integration of the above equation is computed by,
\begin{align}
     \Gamma_{i, j}(\mathbf{x}) = (x_i - \hat{x}_i)(x_j - \hat{x}_j) \times \sum_{\ell=1}^k \sum_{p=1}^m \frac{\ell \times p}{k\times m} \times \frac{f(\hat{\mathbf{x}} + \frac{\ell \times p}{k\times m} (\mathbf{x}-\hat{\mathbf{x}}))}{\partial x_i \partial x_j} \times \frac{1}{k\times m}
\end{align}
Theoretical proof demonstrates that the Integrated Hessian satisfies axioms similar to those fulfilled by integrated gradients. 

\textbf{Integrated Directional Gradients (IDG)} \cite{sikdar2021integrated}. \citeauthor{sikdar2021integrated} propose IDG to explain the model output by computing importance scores for groups of words in NLP domain. The groups of words refer to meaningful subsets of input tokens, which are obtained through the constituency parse tree of input sentences. Given a group of words $S$, the IDG score is calculated by integrating the gradients with respect to the group of features along the straight line from the baseline $\hat{x}$ to the input $x$, shown as follows,
\begin{align}
    \text{IDG($S$)} = \int_{\alpha=0}^1 \bigtriangledown_S f(\hat{\mathbf{x}} + \alpha (\mathbf{x}-\hat{\mathbf{x}})) \: d\alpha
\end{align}
Where $\bigtriangledown_S f = \bigtriangledown f \cdot \hat{z}^s$ and 
\begin{equation}
\hat{z}^s = \frac{z^s}{||z^s||}  
\end{equation}
\begin{equation}
  z_i^s =
    \begin{cases}
      x_i - \hat{x}_i & \text{if } x_i \in S \\
      0 & \text{otherwise}
    \end{cases}       
\end{equation}
Then a dividend $d(S)$ is calculated by normalizing the absolute values of $\text{IDG}(S)$ over all meaningful subsets.
Finally, the importance score of a feature group $S$ is calculated by adding up the ``dividends'' of all the subsets of $S$, i.e., $\{T|T \subseteq S\}$ , including $S$ itself.

\textbf{Guided Integrated Gradients (Guided IG)} \cite{kapishnikov2021guided}. \citeauthor{kapishnikov2021guided} demonstrated that noisy saliency maps are caused by the following factors: (a) the high curvature of a model's decision surface (b) the reference point and integral path and (c) approximation of Riemann integration. The authors observe that for an interpolated image $\mathbf{x}_i$ in saturated regions (e.g., $0.3 < \alpha < 1$), the magnitude of gradients w.r.t. the input can be high even if the multiplication of gradients and directional derivatives contributes little to the prediction differences, i.e., $\frac{\partial f(\mathbf{x}_i)}{\partial \mathbf{x}_i} \times \frac{(\mathbf{x} - \hat{\mathbf{x}})}{m} \ll f(\mathbf{x}) - f(\hat{\mathbf{x}})$ in Eqn.\eqref{eq:dig}. In other words, spurious pixels that have high gradients could accumulate non-zero attributions in the path integral. To avoid gradient integration in nearby points with high curvature , guided IG (GIG for abbreviation) introduces an adaptive integral path that has the lowest absolute value of partial derivatives from the baseline to input. Specifically, the adaptive path is identified by minimizing the overall absolute gradients for all segments from all possible paths,
\begin{align}
    \gamma^{GIG} = \arg\min_{\gamma \in \Gamma}  \sum_{i=1}^N \int_{\alpha=0}^1 |\frac{\partial f(\gamma(\alpha))}{\partial \gamma_i(\alpha)} \frac{\partial \gamma_i(\alpha)}{\partial \alpha}| d\alpha
\end{align}
where $\Gamma$ contains all possible path from the baseline to input $\mathbf{x}$. After determining the optimal path $\gamma_{GIG}$, guided IG accumulates the gradients along the path, similar to IG,
\begin{align}
    g^{GIG}_i (\mathbf{x}) = \int_{0}^1 \frac{\partial f(\gamma^{GIG}(\alpha))}{\partial \gamma^{GIG}_i(\alpha)} \frac{\partial \gamma^{GIG}_i(\alpha)} {\partial \alpha} d\alpha
\end{align}
The search for the optimal path $\gamma^{GIG}$ is impossible without knowing the $\Gamma$ in advance. As such, \citeauthor{kapishnikov2021guided} proposed a greedy strategy to search the best point with the lowest absolute values of partial derivative step by step. At each step, starting from a baseline image, it finds a subset $\mathbb{S}$ (e.g., 10\%) of features whose absolute partial derivatives are lowest, then restores pixels in $\mathbb{S}$ to their corresponding values in the input image, and stops until all pixel intensity are same to the input. 

\textbf{Adversarial Gradient Integration (AGI)} \cite{pan2021explaining}. 
Considering (1) IG may produce inconsistent explanations with different baselines; (2)  the choice of uninformative baseline lacks justification in certain tasks, AGI introduces a baseline-free method. In particular, AGI adopts the gradient descending to search an adversarial example of a class $j$ (different from the current prediction) along the curve of steepest descent, and accumulates the gradients in the descending steps. 
\begin{align}
    g_i^{AGI}(\mathbf{x}) = \int_{til\ adv} -\bigtriangledown_{\mathbf{x}_i}f(\mathbf{x}) \cdot \frac{\bigtriangledown_{\mathbf{x}_i}f_j(\mathbf{x})}{|\bigtriangledown_{\mathbf{x}_i}f_j(\mathbf{x})|}d\alpha
\end{align}
``$til\ adv$'' means it stops until an adversarial example is found or reaches the predefined maximum step. The number of non-target classes may be huge, and then AGI averages the attribution score over multiple randomly selected classes.

\textbf{Boundary-based Integrated Gradients (BIG)} \cite{wang2022robust}. \citeauthor{wang2022robust} theoretically and empirically conclude that gradients based explanations lack robustness because input gradients deviate from the normal vector of nearby decision boundary. For the commonly used ReLU networks, the feature space is divided into many polytopes depending on the activation patterns of ReLUs, and the decision boundaries only contain a set of hyperplane segments of certain polytopes. Given an input $\mathbf{x}$, assuming it lies within the polytope $P$, the model can be expressed as a local linear model $f(x) = w_P \mathbf{x} + b_P$. The gradients $w_P$ may fail to explain the local behavior of model decision process if the polytope is far from the decision boundary. Consequently, the authors introduce the closest adversarial example $\mathbf{x}'$ of $\mathbf{x}$ to facilitate the explanation of $f(\mathbf{x})$. BIG is defined by substituting the baseline of Integrated Gradients (IG) with the adversarial example and subsequently accumulating gradients in a manner similar to IG.
\begin{align}
    g^{BIG}_i(\mathbf{x}) = g^{IG}_i(\mathbf{x}, \mathbf{x}')
\end{align}
where $\mathbf{x}'$ is the closest adversarial example of $\mathbf{x}$ and $\vee\mathbf{x}_m, ||\mathbf{x}_m - \mathbf{x}|| < ||\mathbf{x}' - \mathbf{x}|| \to F(\mathbf{x}) = F(\mathbf{x}_m)$. Furthermore, the gradients of $\mathbf{x}'$ can be employed to augment the conventional vanilla gradients based explanation, termed the Boundary-based Saliency Map (\textbf{BSM}).

\textbf{Important Direction Gradient Integration (IDGI)} \cite{yang2023idgi}. A common challenge in integrated gradients based methods is the presence of noisy explanations during integration. \citeauthor{yang2023idgi} point out one reason is that the multiplication between the gradient and the path segment in each step of Riemann integration. The gradient reflects the steepest direction of the model prediction. However, the path segment can be decomposed by the noise direction and important direction. The multiplication between the noise direction and gradient has no effect on the model prediction but introduces noise into the saliency map. To address this issue, they propose the IDGI, which computes the feature attribution score in each line segment from $\mathbf{x}_j$ to $\mathbf{x}_{j+1}$ by,
\begin{align}
    g^{IDGI}_{i,j} &= \frac{h_i \times h_i \times d}{h \cdot h}, \\
    h &= \frac{\partial f(\mathbf{x}_j)}{\partial x}, \\
    d &= f(\mathbf{x}_{j+1}) - f(\mathbf{x}_{j})
    \label{eq:idgi}
\end{align}
where $g^{IDGI}_{i,j}$ is the importance score of feature $i$ at the step $j$. IDGI directly disregards the multiplication between the noise direction and gradient, focusing solely on the important direction that aligns with the gradient. Summation of Eqn \eqref{eq:idgi} over all segments in a path will obtain the final attribution score. 

\textbf{Negative Flux Aggregation (NeFLAG)} \cite{DBLP:conf/ijcai/0081PLQZ23}. \citeauthor{DBLP:conf/ijcai/0081PLQZ23} reformulates the gradient accumulation via divergence and flux in vector field. Mathematically, the gradient of model prediction w.r.t input instances form the vector field. IG \cite{sundararajan2017axiomatic} accumulates gradients along a single line while AGI \cite{pan2021explaining} aggregates gradients over multiple integral paths to the input, resulting in improved performance. An extended idea is to accumulate all possible paths, which motivates the analysis from the perspective of divergence and flux. Let $\mathbf{F} = \bigtriangledown_x f$ be the vector field, the divergence can be defined by the limit of the ratio of flux through an infinitesimal surface enclosure $S$ that encloses $\mathbf{x}$ to the volume $V$,
\begin{align}
    \text{div } \mathbf{F}|_x = \bigtriangledown \cdot \mathbf{F} = \lim_{V \to 0} \frac{1}{|V|} \oiint\limits_{S(V)} \mathbf{F}\cdot \mathbf{\hat{n}} dS
\end{align}
where $S(V)$ is the boundary of $V$, and $\mathbf {\hat {n}}$ is the outward unit normal of $S(V)$. To interpret a model prediction, NeFLAG integrates the divergence in the neighborhood $V(\mathbf{x})$ of the input $\mathbf{x}$. According to the Divergence theorem,
\begin{align}
    \iiint_V(\bigtriangledown \cdot \mathbf{F})dV = \oiint\limits_{S} (\mathbf{F} \cdot \mathbf{\hat{n}}) dS
\end{align}
The total divergence in the volume $V(\mathbf{x})$ can be converted by calculating the flux through the surface $S$. As the goal is to explain why the model made a particular prediction, they focus on the negative direction where the current prediction decreases due to the absence of certain features. Finally, NeFLAG is defined as,
\begin{align}
    \hat{w} = \oiint_{S_x^-}(\mathbf{F} \odot \mathbf{\hat{n}}) dS
\end{align}
where $S_x^-$ is a set of points $\mathbf{\tilde{x}}$ on the $\epsilon$-sphere surface centered at $\mathbf{x}$ where the flux is negative, and $\hat{w}$ is the feature importance vector. At last, sampled points $\mathbf{\tilde{x}}$ are found using the gradient descent algorithm from $\mathbf{x}$, similar to AGI \cite{pan2021explaining}. 

\subsubsection{Summary} An evident drawback of the aforementioned methods is the presence of noise in irrelevant features and researchers \cite{miglani2020investigating, kapishnikov2021guided} posit that the noise primarily stems from the inadequacies in method design rather than the model inherently relying on noise for predictions. Therefore, the main theme of integrated gradients based studies is to identify potential sources of noise and propose effective solutions to address them., which include: (1): the choice of integral path (BlurIG \cite{xu2020attribution}, Expected IG \cite{sturmfels2020visualizing} Split IG \cite{miglani2020investigating}, NeFLAG \cite{DBLP:conf/ijcai/0081PLQZ23}); (2) the high curvature of a model's decision surface (Guided IG \cite{kapishnikov2021guided}, BIG \cite{wang2022robust}), a.k.a, discontinuous gradients; (3) the Riemann approximation at each step (AGI \cite{pan2021explaining}, IDGI \cite{yang2023idgi}).  Nevertheless, visual noise still persists, prompting us to propose more effective approaches in future endeavors.

\subsection{Bias Gradients based Explanation}
Vanilla gradients and integrated gradients have garnered significant attention in understanding model behaviors, 
however,  it is imperative to note that the bias term also plays a pivotal role in shaping the final output, and its influence is intricately determined by the activation patterns of ReLUs.

For an input $\mathbf{x}$, the DNN in Eqn \eqref{eq:model} can be written as a piecewise-linear model \cite{wang2019bias},
\begin{align}
    f(\mathbf{x}) = \prod_{i=1}^mW^i_\textbf{x} + (\sum_{j=2}^m \prod_{i=j}^m W^i_\textbf{x} b^{j-1}_\textbf{x} + b^m)
\end{align}
where $W^i_\textbf{x}$ is modified from $W^i$ by zeroing out certain rows, and $b^i_{\textbf{x}}$ is a scaled version of $b^i$. $W^i_\textbf{x}$ and $b^i_\textbf{x}$ depend on the activation pattern $\phi^i(W^i z^i + b^i)$ at the layer $i$ ($z^i$ is the input at layer $i$). In particular, 
\begin{align}
    W^i_\textbf{x}[p] &= c^{\phi^i(W^i z^i + b^i)[p]} \cdot W^i\\
    b^i_\textbf{x}[p] &= c^{\phi^i(W^i z^i + b^i)[p]} \cdot b^i
\end{align}
If $\phi(\cdot)$ is the ReLU activation function, the $c$ is the indicator function to set the output of node $p$ to $0$ if not activated. In DNN models, it has been observed that sparse representations are acquired to facilitate straightforward discrimination by the classifier \cite{glorot2011deep}. This observation suggests that within deep layers, fewer ReLUs are activated, consequently leading to diminished gradient magnitudes owing to the multiplication of a series of indicator functions. This phenomenon is formally referred to as ``gradient saturation'' where $b^{i}_\textbf{x}$ substantially contributes to the final output. In \cite{wang2019bias}, they observe that only using the bias term, the model still achieves more than $30\%$ accuracy on CIFAR-100 test set. Therefore, decomposing the contribution of bias terms to input features should be investigated. 

\textbf{FullGrad} \cite{srinivas2019full}. \citeauthor{srinivas2019full} propose to decompose the output prediction into input gradients and bias gradients simultaneously, 
\begin{align}
    f(\textbf{x}; \textbf{b}) = \bigtriangledown_\textbf{x} f(\textbf{x};\textbf{b})^T\textbf{x} + \bigtriangledown_{\textbf{b}}f(\textbf{x};\textbf{b})^T\textbf{b}
    \label{eq:fullgrad}
\end{align}
The $\bigtriangledown_\textbf{x} f(\textbf{x};\textbf{b})$ is the gradient w.r.t. the input $\mathbf{x}$, which is the same to previous methods \cite{simonyan2013deep}. The $\bigtriangledown_{\textbf{b}}f(\textbf{x};\textbf{b})$ is the gradient w.r.t. the bias terms, including the explicit biases in convolutional and fully connected layers, and implicit biases in batch normalization layers. 
In a CNN, the bias parameters exhibit an identical spatial structure as the feature map  due to the weight sharing and sliding window mechanism employed during convolution. Finally, we can visualize the bias contribution $\bigtriangledown_{\textbf{b}}f(\textbf{x};\textbf{b})^T\textbf{b}$ by the saliency map after the post-processing, which is computed by,
\begin{align}
    S(\mathbf{x}) = \psi(\bigtriangledown_\textbf{x} f(\textbf{x};\textbf{b})^T\textbf{x}) + \sum_{l\in L} \sum_{c\in c_l} \psi([\bigtriangledown_{\textbf{b}}f(\textbf{x};\textbf{b})^T\textbf{b}]_c^l)
\end{align}
Here $L$ denotes the number of convolutional layers, $c_l$ is the number of channels at layer $l$, and $\psi(\cdot)$ is a series of post-processing operations, e.g., upsampling, abstract, and min-max normalization. 
Due to the down sampling in the forward pass, it becomes necessary to upsample $\bigtriangledown_{\textbf{b}}f(\textbf{x};\textbf{b})^T\textbf{b}$ to match the image size at each layer $l$. 
The absolute operator and min-max normalization procedures are employed, aligning with other visualization techniques.
While FullGrad theoretically satisfies a set of desired properties, such as weak dependence and completeness, the practical implementation introduces a deviation by discarding the bias gradient in the fully connected layer and incorporating the upsampling operation. In addition, the adaptation of FullGrad to tabular and NLP tasks remained unknown. 

\subsection{Postprocessing for Denoising}
\label{smoothing}
Visual noise is a prevalent issue in gradients based explanations. Here, we introduce two popular techniques to denoise derived explanations through postprocessing.

\textbf{SmoothGrad} \cite{smilkov2017smoothgrad}. \citeauthor{smilkov2017smoothgrad} observes that in the vicinity of the input image, the derivative of the model prediction w.r.t. input fluctuates sharply, even though the natural images are visually indistinguishable from humans. They highlight that noise may emerge due to high local variations of gradients. Consequently, they propose a smoothing method that averages the gradients over samples in the vicinity of an input $\mathbf{x}$, expressed mathematically as:
\begin{align}
    \hat{g}_s(f, \mathbf{x}) = \frac{1}{N} \sum_{i=1}^N g(f, \mathbf{x}+ \delta_i)
    \label{eq:smoothgrad}
\end{align}
where $\hat{g}_s$ represents the smoothed explanation, $N$ is the number of samples, and $\delta_i  \sim \mathcal{N}(0, \sigma^2)$ is the Gaussian noise exerting on the input with standard deviation $\sigma$. 

\textbf{VarGrad} \cite{adebayo2018sanity}. Similar to SmoothGrad, \citeauthor{adebayo2018sanity} propose to compute the variance of explanations for samples in the vicinity of the input image for denoising,

\begin{align}
    \hat{g}_v(f, \mathbf{x}) = \frac{1}{N} \sum_{i=1}^N [g(f, \mathbf{x}+ \delta_i) - \hat{g}_s(f, \mathbf{x})]^2
    \label{eq:vargrad}
\end{align}

While both denoising methods can enhance gradients based explanations, \citeauthor{seo2018noise} \cite{seo2018noise} conclude that SmoothGrad does not smooth the gradient of the prediction function, and VarGrad captures higher-order partial derivatives rather than depends on the gradient of the prediction function. 

\subsection{Summary}
\begin{figure}
    \centering
    \includegraphics[width=0.85\linewidth]{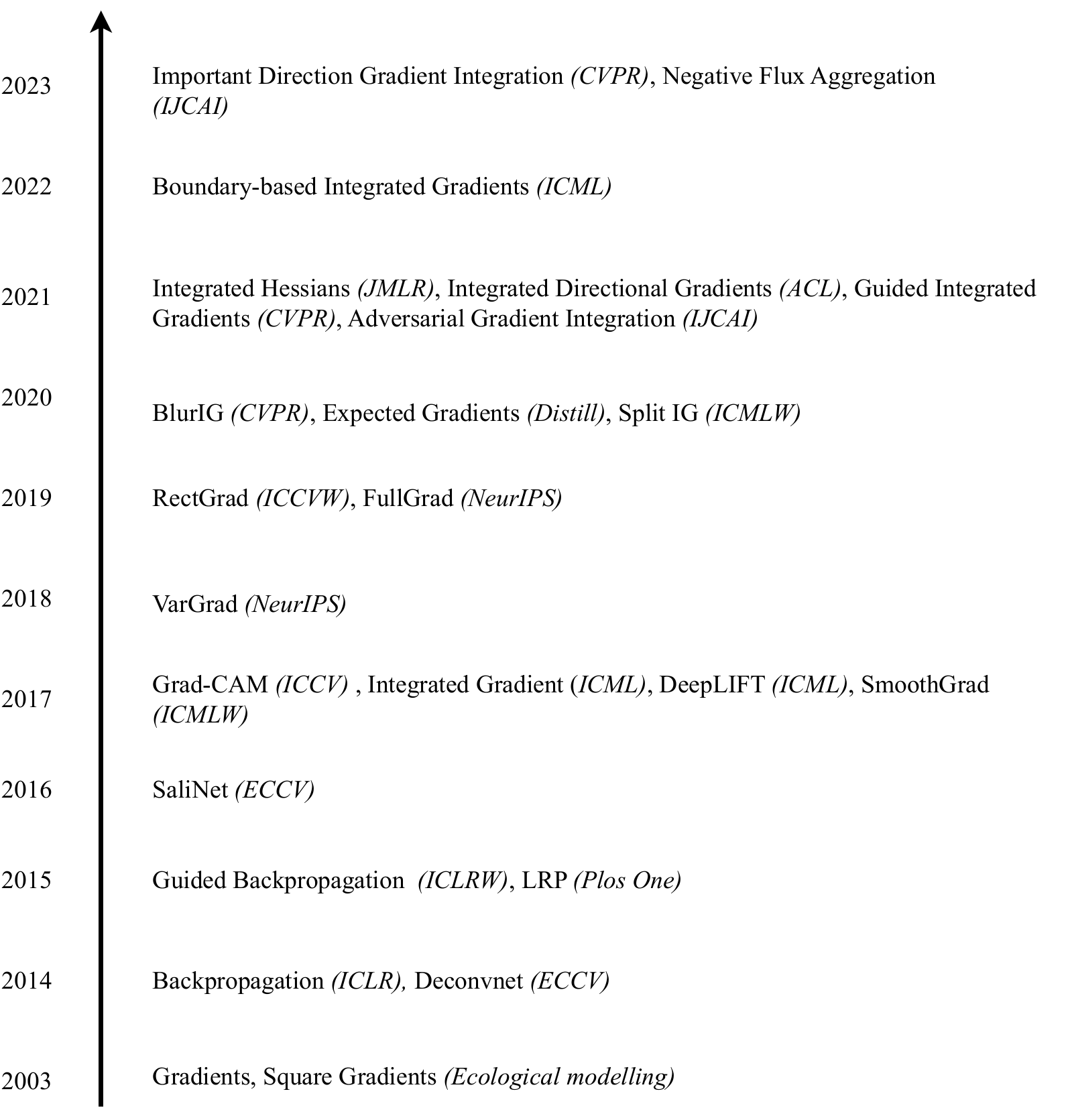}
    \caption{The chronological evolution of gradients based explanations is depicted in the publication timeline.}
    \label{fig:summary}
\end{figure}

We summarize the publication timeline of gradients based explanations in Figure \ref{fig:summary}. 
The timeline provides the publication years, aiding in understanding how each algorithm addresses issues of previous methods and introduces novelty in chronological order. It may help researchers comprehend the latest advancements and cutting-edge developments, thereby enabling them to contribute further improvements to the field.

\section{Evaluation metrics}
\label{sec:evaluation}

Researchers usually evaluate derived explanations from two aspects: explainability and faithfulness. Explainability refers to the capability of making model decisions understandable to humans. In particular, human assessment, fine-grain image recognition, and localization tests are adopted to determine whether the explanations align with user expectations. Faithfulness refers to the extent to which explanations accurately reflect the internal decision-making processes. Specifically, various ablation tests are used to evaluate faithfulness from the causal perspective. Additionally, randomization tests are employed to assess whether explanations are dependent on model parameters and input instances. Regarding other metrics, e.g., algorithm efficiency and interactivity, which are tailored to specific research problems in XAI, we omit their details from our paper.

\subsection{Human Evaluation} 

Human evaluation is devised on the premise that when explanations from a method align with human domain knowledge and relevant experiences, users are more likely to readily accept and endorse this method \cite{Ribeiro2016,selvaraju2017grad}. Therefore, researchers recruit human participants to provide scores on designated scales \cite{kim2018interpretability} or answer designed questions \cite{narayanan2018humans,selvaraju2017grad} after reviewing given explanations. These subjective scores are reported as indicators of human acceptance for comparing different methods. 

Human evaluations have inherent drawbacks. Firstly, they are typically time-consuming and expensive. For example, the median hourly wage of Amazon Mechanical Turk was approximately 2 US dollars in 2017 \cite{10.1145/3173574.3174023}. Secondly, the reproducibility of human evaluation is problematic due to the variability introduced by different participants in separate experiments. Another challenge is the need for specialized domain knowledge in certain evaluations. For instance, in the explanation questionnaire on bullmastiff (an English breed of guard dog) recognition, participants not well-versed in zoology may assign a higher score when an explanation merely emphasizes areas associated with general dogs, rather than highlighting distinguishable features specific to a Bullmastiff.  Crucially, it is imperative to contemplate whether explanations should necessarily conform to human intuition.

\subsection{Localization Tests}

For object detection tasks in computer vision, such as PASCAL VOC \cite{Everingham15} and COCO \cite{lin2014microsoft}, where bounding boxes are annotated, location annotation can be utilized to evaluate whether saliency maps genuinely highlight the target objects.

\textbf{Pointing game} \cite{zhang2018top,qi2019visualizing}. In the pointing game, the maximum importance point is computed over the saliency map. A hit is recorded if the maximum point falls within the bounding box of the correctly predicted category; otherwise, a miss is counted. The pointing game accuracy is computed by the ratio of hits, i.e., \#hits / (\#hits + \#misses).  The overall performance is reported by averaging all scores over different categories on all test samples. 
In an extended pointing game \cite{gu2019understanding}, a thresholding strategy is employed to identify foreground areas covering $p\%$ of the total energy, where the total energy is the sum of all importance values in the saliency map. 
A hit is recorded if the foreground is within the bounding box annotation. A curve is plotted where the $x$ axis represents $p$ and the $y$ axis represents the pointing game accuracy. The Area Under the Curve (AUC) score is reported to measure the localization performance. In addition, \citeauthor{Wang_2020_CVPR_Workshops} \cite{Wang_2020_CVPR_Workshops} introduce the energy-based pointing game that is the total energy within the target bounding box divided by the total energy in the saliency map.

\textbf{Weakly supervised localization} \cite{zhou2016learning}. 
Given top-1 or top-5 predicted classes, we first generate the saliency map of each predicted class. Next, we find segments of regions whose importance value is above 20\% of the max importance value with thresholding methods and then draw the bounding box that covers the largest connected component in each segment. With bounding boxes of a saliency map, localization error can be reported by calculating the IOU (intersection over union) with the ground truth. Finally, the average of location errors over the test dataset is reported to gauge the localization ability of an explanation method. 

\textbf{Fine-grain Image Recognition} \cite{zhou2016learning}. Similar to the weakly supervised localization evaluation, \citeauthor{zhou2016learning} extracts the features within the bounding boxes and leverages the extracted features to train a linear SVM model. Compared with another model trained on the full images' feature, the improved accuracy on the test set can validate the localization ability because the explanation method focuses on the areas of target objects and excludes irrelevant backgrounds. 

\subsection{Ablation Tests}

The localization test may fail to identify the reasons behind a model's specific prediction when explanation methods merely recover input images from output \cite{adebayo2018sanity}. For example, if a model erroneously classifies a husky as a wolf due to the snow in the background, an explanation method, e.g., Guided Backpropagation \cite{springenberg2014striving} and Deconvnet \cite{zeiler2014visualizing}, can still highlight the region of the husky due to partial input recovery. This means the model could incorrectly pass the localization test, attributing the misclassification to the husky rather than the actual cause, the snow in the background.
Therefore, researchers proposed the following evaluation metrics from causal perspectives. The intuition behind these metrics is drawn from conservation and deletion strategies. 
According to \citeauthor{zeiler2014visualizing} \cite{zeiler2014visualizing}, preserving important features should sustain the model's prediction, whereas removing them should significantly decrease the prediction accuracy.


\textbf{Average Drop (AD) and Average Drop in Deletion (ADD)}. \citeauthor{jung2021towards} \cite{jung2021towards} proposed the average drop (AD) and the average drop in deletion (ADD), defined as follows. 
\begin{equation}
    \text{AD} = \frac{1}{N} \sum_{i=1}^N \frac{\max(0, f_c(\mathbf{x}) - f_c(\mathbf{x}_r))}{f_c(\mathbf{x})} \times 100
    \label{eq:ad}
\end{equation}

\begin{equation}
    \text{ADD} = \frac{1}{N} \sum_{i=1}^N \frac{f_c(\mathbf{x}) - f_c(\mathbf{x}_d)}{f_c(\mathbf{x})} \times 100
    \label{eq:add}
\end{equation}
where $f_c(\cdot)$ is the probability of predicted target $c$, $\mathbf{x}_r = \mathbf{a} \odot \mathbf{x}$ denotes the reservation of important regions, and $\mathbf{x}_d = (1 - \mathbf{a}) \odot \mathbf{x}$ denotes the removal of important regions. Here, $\odot$ refers to the Hadamard product. A lower AD and ADD are better.  Similarly, \citeauthor{jung2021towards} also defined the Increase in Confidence (IC):
\begin{equation}
    \text{IC} = \frac{1}{N} \sum_{i=1}^N 1_{[f_c(\mathbf{x}) < f_c(\mathbf{x}_r)]} \times 100
    \label{eq:ic}
\end{equation}
which measures the increase in probability if we remove the irrelevant background. A higher IC is better. Nevertheless, we note that AD and ADD tend to favor explanations highlighting large areas of objective interests. We suggest considering the insertion/deletion score or ABPC below for a more fair evaluation.

\textbf{Insertion/Deletion Score} \cite{petsiuk2018rise,kapishnikov2019xrai}. In a manner akin to AD and ADD, insertion and deletion scores depict the performance increase or decrease by progressively restoring or removing features in the order of importance.  A probability curve can be plotted as a function of the number of features inserted/removed. In this regard,  AD and ADD can be viewed as a specific point on the curve. 

Insertion score (a.k.a., Performance Information Curves (PICs) \cite{kapishnikov2019xrai}). The process begins with an uninformative image, such as a Gaussian-blurred input image without any information relevant to the target prediction. It then sequentially adds back the top-$k$ most important features of the input, as determined by an explanation method, and collects the model prediction for the inserted image. It repeats this procedure until all content of the original image is restored.  The area under the prediction curve (AUC) during insertion process is named the insertion score.  In \cite{kapishnikov2019xrai}, if the y-axis represents average accuracy, they refer to the curve as the Accuracy Information Curve (AIC). When the y-axis is set as the probability of the original prediction, the curve is named the Softmax Information Curve (SIC). A higher insetion score indicates a better explanation.


In deletion score (a.k.a., Area over the Perturbation Curve (AOPC)\cite{bach2015pixel,samek2016evaluating,srinivas2019full}), the goal is to rapidly alter the model's prediction by eliminating important pixels. 
Some researchers \cite{sturmfels2020visualizing,DBLP:conf/ijcai/0081PLQZ23,kapishnikov2019xrai} suggest replacing the important pixels with their Gaussian-blurred counterparts, rather than using a black image \cite{petsiuk2018rise}, to avoid introducing zero values \cite{petsiuk2018rise} that could lead to distribution shift problems. 
The area under the prediction curve (AUC) in successive deletion is termed the deletion score and a lower AUC is better. In AOPC, researchers also propose to perturb pixels based on the Least Relevant First (LeRF) order. In addition, \citeauthor{samek2016evaluating} \cite{samek2016evaluating} introduce the the Area Between Perturbation Curves (ABPC) metric, which calculates the area between two curves that remove the features in top-ranked and bottom-ranked order, respectively.  A higher ABPC score indicates that the explanation method aligns more closely with the model's behavior. 

\textbf{Remove And Retrain (ROAR)} \cite{hooker2019benchmark}. 
The deletion score substitutes the top-$k$ important features with uninformative values and assesses the resulting performance degradation. However, it remains unclear whether the prediction degradation stems from a distribution shift or informative features have been removed. ROAR retrains the model on the substituted data and reports the accuracy drop on the test set, wherein the top-$k$ important features have also been replaced. Retraining helps eliminate potential causes stemming from distribution shifts, allowing a focus on the specific removal of information. Ideally, an accurate explanation method will identify the most important features whose removal will cause a large accuracy drop on the test set. Similar to the deletion curve, we can create a plot where the $x$ axis represents the number of features removed, and the $y$ axis indicates the performance drop. A lower prediction curve on the plot signifies a more effective explanation. However, the drawback of ROAR is the computational burden during model retraining especially when successively perturbing top-$k$ important features. 

\subsection{Randomization Tests}

\citeauthor{adebayo2018sanity} \cite{adebayo2018sanity} observed that some saliency methods (e.g., Guided backpropagation \cite{springenberg2014striving}, Guided Grad-CAM \cite{selvaraju2017grad}) are independent of model parameters and data the model trained on. Despite their superior performance on aforementioned metrics, these explanation methods cannot effectively assist users in understanding model behaviors. To address this limitation, they propose 
two randomization tests as outlined below.
  
\textbf{Model parameter randomization test}. This test compares saliency maps from a trained model with those from a model with the same architecture but initialized randomly.
For example, we can randomly initialize the weights, or shuffle the trained weights of a layer, and compare the differences in saliency maps layer by layer. If the saliency maps are insensitive to model parameters, the explanation method would not help to debug the model. 

\textbf{Data randomization test}. 
This test compares saliency maps from models trained on the original dataset with those from models trained on the same dataset but with shuffled annotated labels. The saliency maps should have significant differences if the explanation method depends on data annotation.  


To quantitatively compare saliency maps before and after the randomization test, the authors utilized Spearman rank correlation, structural similarity index (SSIM), and Pearson correlation of the histogram of gradients (HoGs).

\section{Research Challenges}
\label{sec:challenges}

After describing the algorithm details and evaluations, our discussion shifts towards the research challenges. Initially, we present general challenges inhabited in all feature attribution methods, followed by a detailed exploration of specific concerns within gradient based explanations.

\subsection{General Challenges}

The general challenges arise from all feature attribution problems, and addressing these challenges will also enhance gradient based explanations.

\textbf{Evaluation issues}. The major challenge is to evaluate the correctness of explanations stemming from the following factors: (1) no ground truth. 
The black-box nature of deep models poses challenges in establishing a ground truth for explanations, and even with human annotation, uncertainties persist regarding whether the model's functioning aligns with human understanding. (2) trade-off between incomparable metrics when evaluating explanations from different perspectives. For instance, an explanation method might exhibit high fidelity to the model, but this could come at the expense of explainability. (3) The lack of evaluation tasks. Explanation methods are usually proposed on the model-agnostic assumption and could be generalized to other instances in broad domains. Annotating such a general dataset appears to be an impossible task. 


\textbf{Algorithmic efficiency}. Feature attribution methods typically run on a per-instance basis. When dealing with large datasets containing millions of samples or applications with billions of user requests, generating and analyzing explanations for each instance imposes a significant computational burden.
In a recent study, \citeauthor{wang2023summarizing} \cite{wang2023summarizing} observed that similar users tend to share common importance values and they aggregate feature importance vectors of a dataset at a group level and only pick up top-$k$ important features for efficient summarization. For an explanation system, we believe that more attention should be paid to providing real-time explanations, and efficiently examining historical explanations in the background. 

\textbf{Feature correlations}. Existing feature attribution methods often implicitly assume that features are independent of each other., e.g., LIME \cite{Ribeiro2016} and IG \cite{sundararajan2017axiomatic}. 
However, spatial structures in images pose a challenge when assigning importance scores to each 
feature. For example, a dog in an image remains distinguishable even if we remove just one or two pixels in its body. The model may make incorrect predictions only when significant portions, like the head and body, are removed. Therefore, we should pay attention to the feature interaction and correlation during the explanation. Probably, we should consider all pixels that appear in the regions of the target as a whole or disentangle correlated features for future directions

\textbf{Personalised XAI}. Feature attribution methods are often considered complete upon delivering explanations to end users \cite{lakkaraju2022rethinking, Abdul2018trends, rohlfing2020explanation, zhang2023may}. However, given diverse user backgrounds, these one-size-fits-all explanations may not fulfill all users' intentions. For example, medical practitioners may require explanations to justify the model's decisions, whereas data scientists might be more interested in understanding the strengths and limitations of the model. This discrepancy underscores the necessity for personalized explanations that align with individual user needs. Nevertheless, creating such personalized XAI presents significant challenges. First, First, it requires a deep and continuous analysis of users' backgrounds, expertise, and objectives to accurately identify their specific requirements. Besides, the trade-off between AI transparency and the explanations' complexity should be well-balanced. Overly complex explanations might confuse users rather than enlighten them, especially for those without an AI technical background. It's important to find ways to guide users based on their knowledge, enabling them to gradually comprehend the explanations.
Therefore, it is crucial to provide users with a pathway to gradually understand and engage with AI explanations effectively, based on their existing knowledge levels.



\subsection{Specific Challenges}
Next, we introduce the following challenges limited to gradient based explanations. 

\textbf{The role of bias}. Recent studies \cite{zaken2022bitfit, wang2019bias} discussed the significance of biases on model predictions: when the gradient has a small magnitude, the bias term has a larger contribution.  Remarkably, utilizing only the bias term of the local linear model still achieves an accuracy above $30\%$ on the CIFAR-100 test set \cite{wang2019bias}. Bias terms, while independent of input features, exhibit strong connections in the forward pass. We take the ReLU (i.e., $\phi(wx+b)$) as an example: when $wx+b$ is greater than 0, the ReLU is activated, allowing $wx$ and $b$ to propagate to deeper layers. Conversely, when feeding an all-zero image to a CNN model, only a few ReLUs are activated, limiting the impact of biases on the prediction. Hence, a more in-depth investigation into the influence of bias should be carried out.

\textbf{Hyperparameter sensitive}. The attribution scores could depend on the choice of hyperparameters, e.g., the baseline point in IG \cite{sundararajan2017axiomatic}, the steps in IDGI \cite{yang2023idgi}, the sampling volume $V$ in NeFLAG \cite{DBLP:conf/ijcai/0081PLQZ23}. If we select inappropriate hyperparameters, the explanation method may produce inconsistent, even counter-intuitive explanations. However, the hyperparameter tuning is independent of the input instance, merely driven by a higher evaluation score, which lacks clear intuition for better explainability. Investigating the hyperparameter-free explanation method or one with clearly interpretable hyperparameters should be a focus of future research. 

\textbf{Lack rigorous proof}. Numerous recent studies are built on unproven hypotheses. Taking the gradient saturation problem as an example, when the predicted probability is close to $1$, the gradients has a small magnitude, which motivates the series of integrated gradient based methods.  Nevertheless, the theoretical connections between noisy visualization and smaller gradients remain unknown, and the reasons why accumulated gradients can eliminate noise and identify important features lack adequate justification. While numerous integrated gradient based methods satisfy certain desired properties, it is crucial to note that these properties are necessary but insufficient for establishing connections between the attribution scores and the final prediction \cite{doi:10.1073/pnas.2304406120}. Therefore, more attention should be devoted to exploring and discovering both empirical and theoretical evidence to substantiate the hypotheses underlying these methods, rather than directly building algorithms solely based on these hypotheses.  

\textbf{Model security and privacy}. When offering explanations to end users to elucidate the prediction logic, it unavoidably exposes the model to various potential attacks \cite{10.1145/3287560.3287562,10.1145/3531146.3533188}. For instance, in the case of a linear model, the gradients are equivalent to the model weights. If we provide gradients as explanations, it's akin to disclosing the internal structure of the model. In \cite{10.1145/3287560.3287562}, \citeauthor{10.1145/3287560.3287562} theoretically and empirically prove that extracting the model from gradients is the order of magnitude more efficient than extracting it from the prediction interface. 
Given the substantial investment in model development, safeguarding models in XAI should receive considerable attention.


\textbf{Fragility of explanations}. 
Recent research \cite{ghorbani2019interpretation,wang2020gradient} demonstrates that explanations are also vulnerable to adversarial attacks. For instance, perturbing certain pixels where there is a perceptually indistinguishable difference in input and model prediction also remains the same, can lead explanation methods to produce dramatically different saliency maps. The fragility of explanations raises significant security concerns for explanation methods especially when attackers manipulate input features to jeopardize the credibility of an explanation method. \citeauthor{ghorbani2019interpretation} points out that the fragility derives from the high dimensionality and non-linearity of neural networks, but addressing this issue remains an unsolved challenge. Therefore, developing robust explanation methods and implementing countermeasures to audit adversarial attacks should be a focal point in deployment.

\section{Conclusion}
\label{sec:conclusion} 

XAI research has seen a surge recently. Despite the abundance of surveys in the XAI field, many of them aim to cover all facets of XAI research, providing only provide only brief introductions to gradient based feature attribution methods with limited depth. A comprehensive overview of algorithmic breakthroughs of gradient based explanations remains largely unexplored. In this paper, we systematically review recent approaches, categorizing gradient based explanations into four groups. Subsequently, we introduce representative algorithms within each group, elucidating their motivation and how they address previous weaknesses. Furthermore, we illustrate commonly used evaluation metrics for comparing various explanations. Most importantly, we discuss several general challenges in feature attribution methods and specific challenges in gradients-based explanations. We believe that our research will provide a comprehensive understanding to AI researchers and further motivate advanced research in the realm of gradient based explanations.

\bibliographystyle{ACM-Reference-Format}
\bibliography{ref}

\end{document}